\documentclass{article}
\usepackage{spconf,amsmath,graphicx}

\usepackage{bm}
\usepackage{multirow}
\usepackage{color,xcolor}
\usepackage{amsmath}
\usepackage{algorithm}
\usepackage{algpseudocode}
\usepackage[pagebackref=true,breaklinks=true,letterpaper=true,colorlinks,bookmarks=false,citecolor=blue]{hyperref} 
\usepackage{graphicx}
\usepackage{arydshln} 
\usepackage{amssymb}
\usepackage{booktabs} 
\usepackage[utf8]{inputenc} 
\usepackage{makecell}

\newcommand\bb[1]{\textbf{#1}}
\newcommand\ul[1]{\underline{#1}}

\title{MULTI-HEAD KNOWLEDGE DISTILLATION FOR MODEL COMPRESSION}
\name{Huan Wang$^{2*}$\thanks{*~Work done when Huan Wang was an intern at MERL.} \qquad Suhas Lohit$^{1\dagger}$\thanks{$^\dagger$~Corresponding author is Suhas Lohit: \texttt{slohit@merl.com}.} \qquad Michael Jones$^1$ \qquad Yun Fu$^{2}$}
\address{$^1$Mitsubishi Electric Research Laboratories, Cambridge, MA, USA \\
        $^2$Northeastern University, Boston, MA, USA}
\begin{document}
%
\maketitle
\begin{abstract}
Several methods of knowledge distillation have been developed for neural network compression. While they all use the KL divergence loss to align the soft outputs of the student model more closely with that of the teacher, the various methods differ in how the intermediate features of the student are encouraged to match those of the teacher. In this paper, we propose a simple-to-implement method using auxiliary classifiers at intermediate layers for matching features, which we refer to as multi-head knowledge distillation (MHKD). We add loss terms for training the student that measure the dissimilarity between student and teacher outputs of the auxiliary classifiers. At the same time, the proposed method also provides a natural way to measure differences at the intermediate layers even though the dimensions of the internal teacher and student features may be different. Through several experiments in image classification on multiple datasets we show that the proposed method outperforms prior relevant approaches presented in the literature. 

\end{abstract}
\section{Introduction}
\label{sec:intro}
In terms of recognition accuracy, the best deep learning models in image classification typically have a very large number of parameters, on the order of several million trainable weights and biases, as well as several dozens of convolutional layers. Such large and deep learning models, even though they are accurate, make them difficult to be deployed directly in memory- and compute-constrained environments such as mobile devices, and other embedded systems. Neural network compression methods try to solve this problem by compressing large models to small ones without seriously compromising the performance. They mainly fall into multiple categories including pruning~\cite{han2015deep,he2017channel,li2017pruning,MolTyrKar17}, factorization~\cite{jaderberg2014speeding,zhang2015efficient}, and knowledge distillation~\cite{hinton2015distilling}. In this paper, we focus on knowledge distillation, which is used to transfer knowledge from a large neural network, called the teacher, to a smaller and/or shallower neural network, called the student, both trained for the same task. We consider the common task of image recognition in this paper -- given an image, the classifier maps the image to one of $K$ predefined categories. The teacher is trained first using the usual cross-entropy loss. For training the student, both the cross-entropy loss and certain knowledge distillation loss terms are employed. The specific distillation loss function is up to the particular distillation method used. The effectiveness of a method can be evaluated in terms of the student performance, applicability to a large class of teacher-student pairs, and ease of implementation. 

In this paper, we propose a novel method of knowledge distillation by employing auxiliary classifiers at intermediate layers of the teacher and student networks, and use KL divergence~\cite{kullback1997information} loss between the auxiliary classifier outputs as additional loss terms for training the student. This method is simple to implement and readily applicable even in cases where the number of feature dimensions of the student and teacher at the intermediate layers are different. As we will show in the experiments, the proposed algorithm outperforms existing related knowledge distillation methods.

\section{Related Work}
\vspace{-0.1in}
\label{sec:related_work}
In this section, we provide a brief review of closely related knowledge distillation methods. For a more comprehensive treatment of neural model compression techniques, we refer the reader to the survey by Cheng et al.~\cite{cheng2017survey}. The earliest ideas in knowledge distillation came from Bucilua et al.~\cite{bucilua2006model} and Hinton et al.~\cite{hinton2015distilling}. They proposed an effective framework for transferring knowledge from a teacher to a student network by using the KL divergence between the teacher and student outputs as an additional loss function for training the student. This is now a standard loss term used in all knowledge distillation methods, and it usually referred to as ``KD" loss, and we will use the same terminology here. The KD loss only takes into account the discrepancy between the \textit{outputs} of the teacher and student. The focus of subsequent research has been to design loss functions that can further exploit the information from \emph{intermediate} layers, in addition to the outputs. We discuss the main works here. Romero et al. designed FitNet~\cite{romero2014fitnets} where the Euclidean distance between the teacher and student features are used as the loss function. In order to match feature sizes, the authors use additional non-linear layers to map both sets of features to the same number of dimensions, which is an additional hyperparameter. Park et al.~proposed Relational Knowledge Distillation (RKD) \cite{park2019relational} where only pairwise relationships computed among teacher and student features separately and batch-wise are encouraged to be similar, rather than the features themselves. Peng et al.~\cite{peng2019correlation} proposed computing correlations among instances for both teacher and student features and using a loss function that minimizes the difference of these correlations between teacher and student. More recently, Tian et al.~\cite{tian2019contrastive} designed an approach using contrastive learning called CRD. Here, for each teacher and student pair corresponding to the same image, known as a ``positive" pair, a large set (tens of thousands) of ``negative" features derived from different images are stored in memory. The loss function encourages the positive pair to come closer while pushing the negative pairs apart. We consider this idea to be complementary to all the other methods such as FitNet and KD, where only the positive pairs are used for training without need for a large memory bank to store the negative features. Also related is the work by Szegedy et al.~\cite{szegedy2015going} who used auxiliary classifiers for the task of image classification. We employ them here in a novel fashion for the task of model compression via KD.

\begin{figure}[t]
    \centering
    \includegraphics[width=\linewidth]{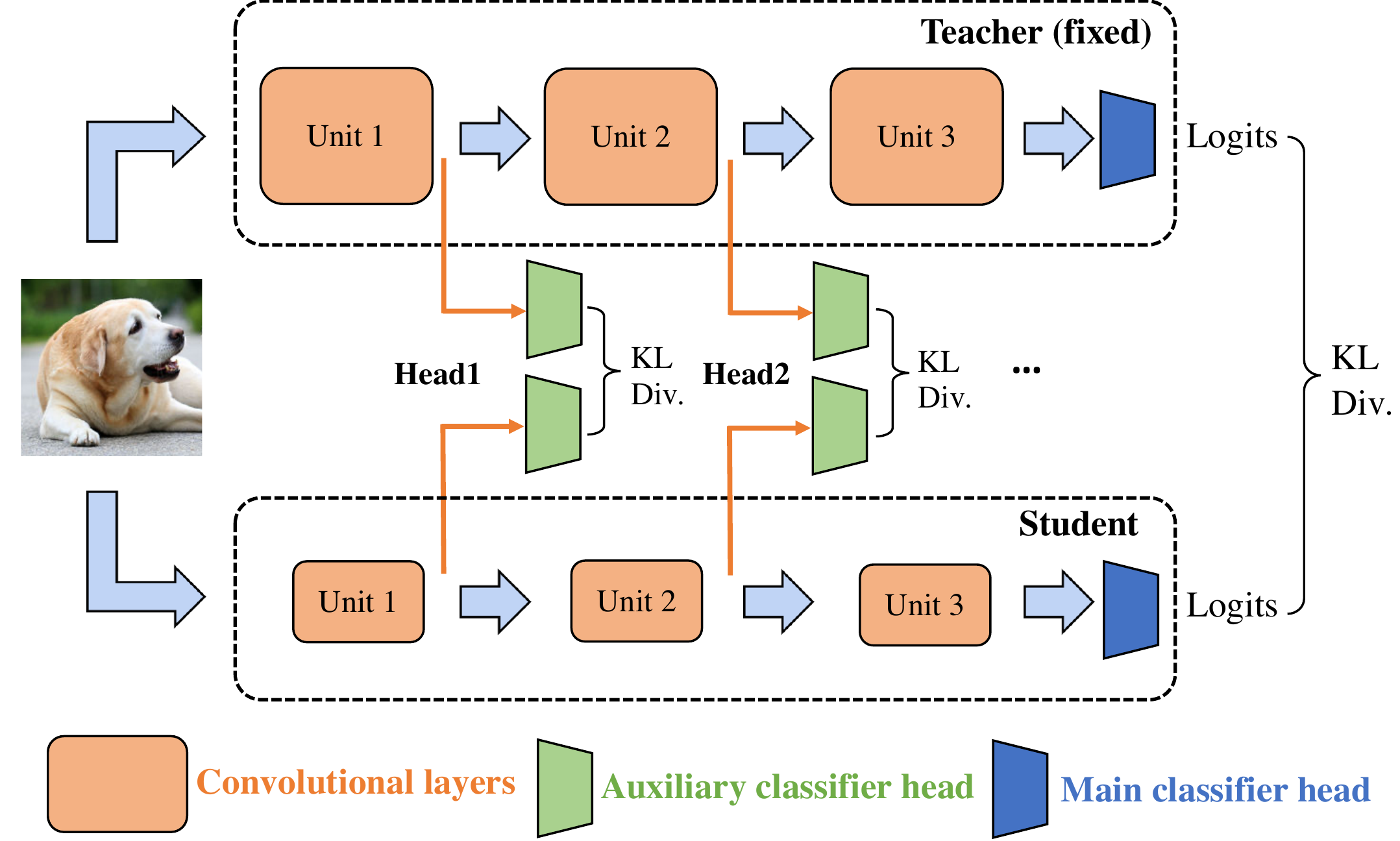}
    \caption{Illustration of the proposed multi-head knowledge (MHKD) method. In addition to the Hinton et al.~KD loss~\cite{hinton2015distilling}, we add auxiliary classifier heads at the intermediate layers of both the teacher and student networks. The KL divergence (KL Div.) between the student and teacher outputs for these auxiliary classifier heads are added as additional loss terms while training the student layers. Note that the auxiliary classifier heads are also trained, however, they will be removed at test time (\emph{best viewed in color}).}
    \label{fig:block_diagram}
\vspace{-1em}
\end{figure}

\section{Multi-head Knowledge Distillation}
\vspace{-0.1in}
\label{sec:method}
Deep convolutional neural networks typically consist of multiple convolutional (conv) units with normalization and non-linear layers followed by a fully-connected classifier. Consider a convolutional layer, parameterized by a 4D tensor $\Theta \in \mathbb{R}^{C_o\times C_i\times kH \times kW}$, where $C_o,C_i,kH,kW$ are the number of output channels, input channels, height, and width of the conv kernel, respectively. It will produce a feature of size $\mathbb{R}^{C_o \times H \times W}$, where $H, W$ are the height and width of the feature. For knowledge distillation in the intermediate layers, the central problem is to build a loss function between the teacher's feature $\mathbf{f}_{T}$ and the student's feature $\mathbf{f}_{S}$. Note that in practice, $\mathbf{f}_{T}$ can be of a different size from $\mathbf{f}_{S}$ because of different network architectures (such as different $C_o,H,W$) and any distillation method for the intermediate layers should overcome this problem.

\vspace{-0.5em}
\subsection{Auxiliary classifier head architecture}
In order to better measure the differences between teacher and student features, we introduce \emph{auxiliary classifier heads}, denoted by $H$. Specifically, for a feature of size $\mathbb{R}^{C_o\times H \times W}$, a small classifier made up of $N_c$ convolutional layers and $N_f$ fully-connected layers is connected to the feature's layer. The output dimension of this small classifier is the same as the final output of the teacher and student networks, namely, the number of classes $K$ and the KL divergence is used as the measure of dissimilarity. For a chosen intermediate layer, we add an auxiliary classifier head for both the teacher and the student which takes in features from that layer and maps them to a probability distribution over $K$ classes. This is illustrated in Figure \ref{fig:block_diagram}. As the parameters of the auxiliary classifier are learned, an immediate dilemma is that a well-trained classifier will make the teacher-student discrepancy much smaller after the mapping even though the true discrepancy can be large before the mapping. Neural networks with enough capacity are universal approximators~\cite{HornikEtAl1989}, therefore, we need to restrict the capacity of the auxiliary classifier. In our experiments, we employ $N_c=2$, $N_f=2$ with $256$ kernels of size $3 \times 3$ in the convolutional layers. The convolutional units contain batch-normalization and the ReLU non-linearity. The proposed additional classifier heads are simple in design yet we will show that the proposed method based on them can outperform the baselines significantly. An important point to note here is that, the teacher and student features are automatically mapped to the same number of dimensions as both the teacher and student classifier heads both map to $K$ classes, and thus, the number of dimensions in the common space is no longer a hyperparameter, unlike the case in FitNet~\cite{romero2014fitnets}.


\begin{table*}[t]
\centering
\caption{Results on CIFAR-100 test set. Our results are obtained using $3$ runs with different random initialization of the parameters with mean and (std) of test set accuracies reported. The best results are in bold and second best underlined. All the results except those of our method are taken from the CRD paper~\cite{tian2019contrastive}. FitNet, CC, RKD and CRD employ the KD loss \cite{hinton2015distilling} as part of the loss function. We can easily see that MHKD outperforms all comparable baselines.}
\setlength\tabcolsep{9pt} 
\vspace{0.5em}
    \begin{tabular}{lcccccc}
    \toprule
    Teacher & WRN-40-2~\cite{zagoruyko2016wide} & ResNet110~\cite{he2016deep} & VGG13~\cite{simonyan2014very} & VGG13 & ResNet32x4   \\
    Student & WRN-16-2 & ResNet20 & VGG8 & MobileNetV2~\cite{sandler2018mobilenetv2} & ShuffleNetV2~\cite{ma2018shufflenet} \\
    Compression (\%) & 68.81 & 70.00 & 58.10 & 92.82 & 83.22\\
    \midrule
    Teacher baseline & 75.61 & 74.31 & 74.64 & 74.64 & 79.42 \\
    Student baseline & 73.26 & 69.06 & 70.36 & 64.60 & 71.82 \\
    \midrule
    KD~\cite{hinton2015distilling} & 74.92 & 70.67 & 72.98 & 67.37 & 74.45 \\
    FitNet~\cite{romero2014fitnets} & \ul{75.12} & 70.67 & \ul{73.22} & 66.90 & \ul{75.15} \\
    CC~\cite{peng2019correlation} & 75.09 & \ul{70.88} & 73.04 & \ul{68.02} & 74.72 \\
    RKD~\cite{park2019relational} & 74.89 & 70.77 & 72.97 & 67.87 & 74.55 \\
    \bb{MHKD (ours)}  & \bb{75.28} (0.26) & \bb{71.00} (0.24) & \bb{73.96} (0.19) & \bb{69.11} (0.12) & \bb{76.34} (0.24) \\
    \midrule
    CRD~\cite{tian2019contrastive} & 75.64 (0.21) & 71.56 (0.16) & 74.29 (0.12) & 69.94 (0.05) & 76.05 (0.09) \\
    CRD+\bb{MHKD} & \bb{75.89} (0.10) & \bb{71.76} (0.10) & \bb{74.37} (0.26) & \bb{70.21} (0.19) & \bb{76.47} (0.21) \\
    \bottomrule
\end{tabular}
\label{tab:result_cifar100}
\vspace{-0.2in}
\end{table*}

\vspace{0.5em}
\noindent \textbf{Loss functions}. (1) The loss function for the teacher's classifier head $H_T$ is simply the cross-entropy between the classifier logits $H_T(\mathbf{f}_T)$ and the one-hot label $\mathbf{y}$.
%
(2) For the student's classifier head $H_S$, the KL divergence between the teacher and student outputs is applied with the cross-entropy loss. This loss form is the same as \cite{hinton2015distilling}, but applied at the intermediate layers using $H_T$ and $H_S$. Then we have the one-head KD (OHKD) version of our proposed algorithm,
\begin{equation}
    \begin{split}
        & \mathcal{L}_{KL} = \mathcal{D}_{KL}(\frac{H_T(\mathbf{f}_T)}{\tau}, \frac{H_S(\mathbf{f}_S)}{\tau}) * \tau^2, \\
        & \mathcal{L}_{OHKD} = \alpha \mathcal{L}_{KL}  + (1-\alpha) \mathcal{L}_{CE}(\mathbf{y}, H_S(\mathbf{f}_S)),
    \end{split}
\label{eq:OHKD}
\end{equation}
where $\mathcal{D}_{KL}$ stands for the KL divergence; $\tau$ is the temperature parameter introduced in the original KD method~\cite{hinton2015distilling} to smooth the probability; $\mathcal{L}_{CE}$ denotes the cross-entropy loss; $\alpha$ is the weight to balance the two loss terms. All the classifier heads are jointly trained with the original KD process. At test time, these auxiliary heads are removed. 

\vspace{-0.1in}
\subsection{Multi-Head Knowledge Distillation (MHKD)}
The above-described auxiliary heads can be repeated at multiple intermediate layers, which gives us the complete proposed \emph{multi-head knowledge distillation} method. Multi-head scheme introduces \emph{deeper} supervision than the one-head counterpart does, which we show to be beneficial to the student learning. When applied at $D$ different layers (see Figure~\ref{fig:block_diagram}), the total loss for training the student is given by
\begin{equation}
    \begin{split}
        \mathcal{L}_{MHKD} &= \beta \sum_{j=1}^{D} \mathcal{L}_{OHKD}^{(j)} + \mathcal{L}_{KD},
    \end{split}
\label{eq:MHKD}
\end{equation}
where $\mathcal{L}_{KD}$ is the original KD loss \cite{hinton2015distilling} (made up of the cross-entropy and KL divergence losses), applied at the \emph{final outputs} of the networks; $\beta$ is a multiplier to balance to two loss terms. How to pick layers to enforce the proposed algorithm can be tricky as there can be hundreds of layers in a modern deep neural network. In practice, such networks are typically designed with multiple units/stages (layers in the same unit have feature maps of the same spatial size) such as VGG~\cite{simonyan2014very}, ResNet~\cite{he2016deep}, and Wide-ResNet~\cite{zagoruyko2016wide}. In our method, the auxiliary classifier heads are added \emph{at the end of these units} (shown in Figure~\ref{fig:block_diagram}) so that they span the whole network evenly.

\section{Experimental Results}
\label{sec:results}
\vspace{-0.1in}
In this section, we evaluate our approach on two real-world datasets CIFAR-100~\cite{KriHin09} and SVHN (Street View House Numbers)~\cite{svhn}. CIFAR-100 consists of $60,000$ $32\times32$ color images of $100$ classes, among which $50,000$ are employed for training and $10,000$ for testing. SVHN consists of over $600,000$ $32\times32$ color images of 10 classes. We employ the standard pre-processing procedures (random crop and horizontal flip) for data-augmentation. We compare our method with popular methods -- FitNet~\cite{romero2014fitnets}, CC~\cite{peng2019correlation}, and RKD~\cite{park2019relational} as they focus on the same topic of bridging intermediate feature size mismatch in KD as we do. The networks we evaluate in this work have at least three units (see Figure~\ref{fig:block_diagram}), so we apply our method at the end of the first three units, i.e., $D = 3$ in Eq.~\ref{eq:MHKD}. For every setting, we randomly run each experiment three times and report the mean and standard deviation of accuracies obtained on the test set. The code and trained models will be made publicly available.

\begin{table*}[h]
\centering
\caption{Comparison of one-head scheme vs.~multi-head scheme in the proposed approach on CIFAR-100. The best results of our method are in bold and second best underlined.}
\setlength\tabcolsep{3pt} 
\vspace{0.5em}
    \begin{tabular}{lcccccccc}
    \toprule
    Teacher/Student & KD~\cite{hinton2015distilling} & FitNet~\cite{romero2014fitnets} & CC~\cite{peng2019correlation} & RKD~\cite{park2019relational} & Head-1 & Head-2 & Head-3 & Head-1+2+3 \\
    \midrule
    WRN-40-2/WRN-16-2 & 74.92 & 75.12 & 75.09 & 74.89 &  \ul{75.27} (0.27) & 75.16 (0.15) & 75.00 (0.35) & \bb{75.28} (0.26) \\
    VGG13/VGG8 & 72.98 & 73.22 & 73.04 & 72.97 & 73.63 (0.29) & \ul{73.76} (0.12) & 73.32 (0.21) & \bb{73.96} (0.19) \\
    \bottomrule
\end{tabular}
\label{tab:ablation}
\vspace{-0.15in}
\end{table*}

\begin{table*}[t]
\centering
\caption{Results on SVHN test set. Our results are obtained using $3$ runs with different random initialization of the parameters with mean and (std) of accuracies reported. The best results are in bold and second best underlined.}
\setlength\tabcolsep{10pt} 
\vspace{0.5em}
    \begin{tabular}{lcccccc}
    \toprule
    Teacher & WRN-40-2~\cite{zagoruyko2016paying} & ResNet56~\cite{he2016deep} & VGG13~\cite{simonyan2014very} & MobileNetV2~\cite{sandler2018mobilenetv2} & ShuffleNetV2~\cite{ma2018shufflenet}   \\
    Student & WRN-16-2 & ResNet20 & VGG8 &  VGG13 & ResNet32x4 \\
    \midrule
    Teacher baseline & 94.29 & 93.69 &  94.20 & 95.47 & 96.08 \\
    Student baseline & 93.05 (0.01) & 93.05 (0.01) & 91.74 (0.06) & 93.96 (0.21) & 94.17 (0.09) \\
    \midrule
    KD~\cite{hinton2015distilling} & 94.67 (0.06) & 93.95 (0.12) & 93.93 (0.06) & 95.31 (0.04) & 95.96 (0.10) \\
    FitNet~\cite{romero2014fitnets} & 94.75 (0.06) & 94.10 (0.05) & 94.25 (0.06) & 95.73 (0.09) & 96.20 (0.08) \\
    CC~\cite{peng2019correlation}     & 95.00 (0.06) & 94.43 (0.09) & 94.30 (0.12) & 95.75 (0.08) & 96.04 (0.09) \\
    RKD~\cite{park2019relational}    & \ul{95.28} (0.12) & \ul{94.57} (0.06) & \ul{94.40} (0.08) & \ul{95.83} (0.02) & \ul{96.25} (0.02) \\
    \bb{MHKD (ours)} & \bb{95.38} (0.03) & \bb{94.70} (0.04) & \bb{94.45} (0.07) & \bb{95.88} (0.01) & \bb{96.68} (0.03) \\
    \bottomrule
\end{tabular}
\label{tab:result_svhn}
\vspace{-0.1in}
\end{table*}

\vspace{-0.1in}
\subsection{CIFAR-100}
\vspace{-0.1in}
We adopt the same hyperparamters for training all the networks -- SGD optimizer with a batch size of $64$, initial learning rate of $0.05$ reduced by $\frac{1}{10}^{th}$ after epochs $150, 180, 210$. The models are trained for a total of $240$ epochs. The temperature $\tau = 4$, loss weight $\alpha=0.9$ (Eq.~\ref{eq:OHKD}), $\beta=0.5$ (Eq.~\ref{eq:MHKD}). We evaluate our approach with five different teacher-student pairs with the student networks having $58-93\%$ fewer weights than the corresponding teacher networks. On two of them, the teacher and student also have different architecture types.

The results are shown in Table~\ref{tab:result_cifar100}. The teacher and student baseline accuracies are obtained with the original cross-entropy training, i.e., no distillation loss is used. First, the proposed MHKD approach consistently outperforms the original KD method~\cite{hinton2015distilling}. This shows the intermediate supervision indeed helps the student training across different networks. Second, compared to the other relevant methods (FitNet~\cite{romero2014fitnets}, CC~\cite{peng2019correlation}, RKD~\cite{park2019relational}), our scheme is more accurate. Notably, for several pairs (e.g., VGG13/MobileNetV2, ResNet32x4/ShuffleNetV2), our MHKD surpasses the counterpart methods by \bb{more than 1\%}. It is worth emphasizing that our approach does not introduce any new knowledge definition as FitNet~\cite{romero2014fitnets}, CC~\cite{peng2019correlation}, and RKD~\cite{park2019relational} do but applies simple KL divergence loss, as in \cite{hinton2015distilling} with the proposed classifier heads for the internal layers, which also automatically resolves the feature dimension mismatch problem. Finally, note that our method can be easily combined with CRD~\cite{tian2019contrastive}, which is an orthogonal method employing negative pairs and a contrastive loss function. This combination delivers the new \emph{state-of-the-art} results as shown in the last row of Table~\ref{tab:result_cifar100}.

\vspace{0.2em}
\noindent \textbf{Ablation Study}.\label{subsec:ablation}
Here we examine the effect of using multiple heads instead of one in our proposed approach. The results are shown in Table~\ref{tab:ablation}. Note that Head-$j$ refers to an auxiliary classifier head added after Unit-$j$ in the network. (1) When using a single head, the best performance appears at different heads for different network architectures: WRN peaks at Head-1 while VGG peaks at Head-2. However, with all $3$ heads combined, we obtain \emph{even better} results without any need to manually pick the best head position. This shows that more supervision along the whole network is indeed beneficial to the student learning. (2) Note that even the \emph{worst} performance obtained by just one auxiliary head is still very competitive (better than RKD in the WRN case and better than all the three comparison methods in the VGG case), justifying the effectiveness of our algorithm. 

\subsection{SVHN}
\vspace{-0.1in}
We now evaluate our method on the SVHN dataset, where the loss weight $\beta$ for the MKHD loss term is set to $5$ by empirical validation. The selected teacher-student pairs are similar to those in the CIFAR-100 case. Here, we use a batch size $200$, initial learning rate $0.001$ reduced by $\frac{1}{10}^{th}$ after epochs $80, 110, 135$. The models are trained for 150 epochs in total. All the other hyperparameters have the same values as in the experiments on CIFAR-100.

The results are shown in Table~\ref{tab:result_svhn}. When we evaluate the proposed method on teacher-student pairs of different architectures (i.e., the last two columns), we empirically find MobileNetV2/ShuffleNetV2 has better accuracies than VGG13/ResNet32x4, which is different from the case on CIFAR-100 (Table~\ref{tab:result_cifar100}), therefore we switch the teacher-student pairs for MobileNetV2/VGG13 and ShuffleNetV2/ResNet32x4 since it is more reasonable for the stronger model to be the teacher. As observed, similar to the case on CIFAR-100, our method is \emph{consistently better} than all the baseline approaches. This shows that our method is robust across different datasets, unlike the methods like FitNet and RKD which exhibit opposite trends on these two datasets.

\section{Conclusion}
\label{sec:conclusion}
\vspace{-0.1in}
We have presented the multi-head knowledge distillation (MHKD) algorithm in this work, which introduces a novel method for teacher's supervision for the student network at intermediate layers for improved knowledge distillation. Unlike prior art which focus on proposing new definitions of knowledge and loss functions, we have proposed adding simple classifier heads at these intermediate layers of the student and teacher. The corresponding outputs of these classifiers are encouraged to be similar by minimizing the KL divergence between them as the original KD method does. Simple as it is, we have empirically shown the proposed method is very effective and robust across different networks and datasets when compared with state-of-the-art methods.

\vfill\pagebreak

\bibliographystyle{IEEEbib}
\bibliography{references}

\end{document}